\pdfoutput=1
\documentclass[10pt,twocolumn,letterpaper]{article}

\usepackage[pagenumbers]{cvpr}
\usepackage{cuted} %

\usepackage{array}
\usepackage{multirow}

\newcommand{\best}[1]{\textbf{#1}}

\newcommand{\tablestd}[1]{{\normalfont{\fontsize{5}{5}\selectfont$\pm$}\kern0.45pt#1}}
\newcommand{\valstd}[2]{{\renewcommand{\arraystretch}{0.85}\begin{tabular}[c]{@{}r@{}}#1\\[-0.6pt]\tablestd{#2}\end{tabular}}}

\definecolor{cvprblue}{rgb}{0.21,0.49,0.74}
\usepackage[pagebackref,breaklinks,colorlinks,allcolors=cvprblue]{hyperref}

\def\confName{CVPR}
\def\confYear{2027}

\title{\texorpdfstring{\raisebox{-0.4\height}{\includegraphics[height=2.3em]{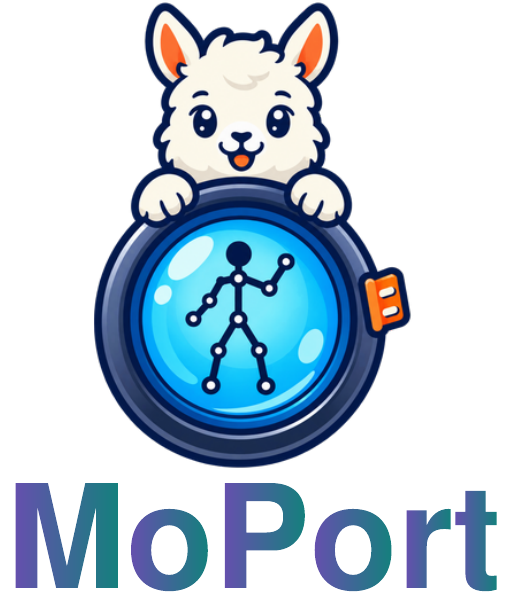}}\,MoPort}{MoPort}: A View-Aware 2D Interface for Motion Language Models}
\author{%
Lixing Tan$^{1}$\qquad Qing Xia$^{1,}$\thanks{Corresponding author: \href{mailto:xiaqing@buaa.edu.cn}{\texttt{xiaqing@buaa.edu.cn}}.}\qquad Yuting Guo$^{2}$\qquad Shuai Li$^{1}$\qquad Aimin Hao$^{1}$\\[3pt]
$^{1}$Beihang University\qquad $^{2}$Beijing Information Science and Technology University\\[3pt]
{\normalsize Project page: \url{https://lextran.github.io/MoPort/}}
}

\begin{document}
\maketitle
\begin{strip}
\centering
\includegraphics[width=\textwidth]{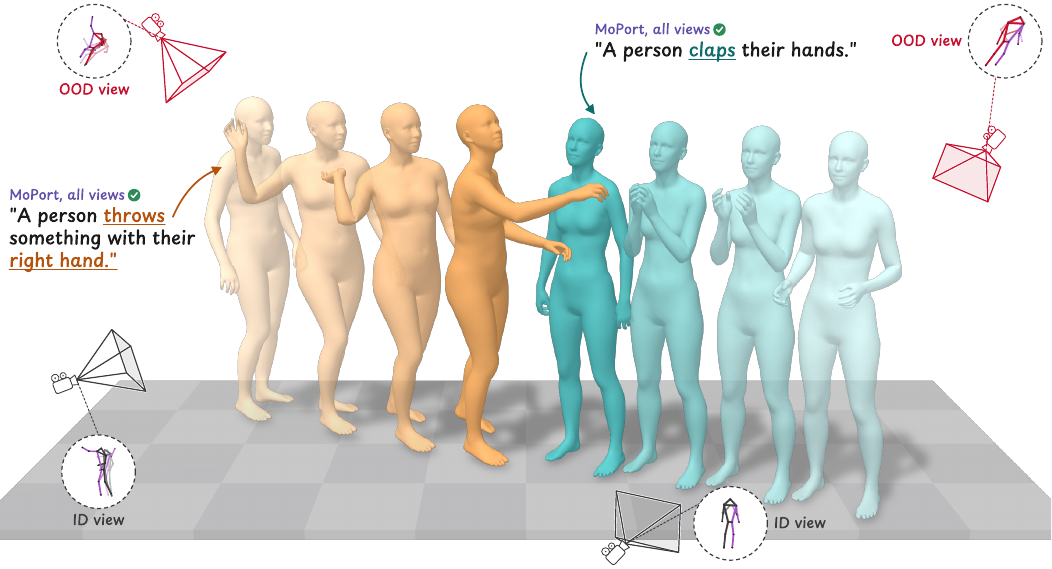}
\captionsetup{hypcap=false}
\captionof{figure}{We introduce \textbf{MoPort}, a view-aware 2D interface
that enables pretrained motion language models to take 2D observations as input.
MoPort gives correct captions under two in-distribution (ID) and two
out-of-distribution (OOD) randomly sampled cameras. Dashed circles show the
2D joints observed by each camera.}
\captionsetup{hypcap=true}
\label{fig:teaser}
\end{strip}

\begin{abstract}
Motion language models (MoLMs) encode 3D human motion as discrete tokens for
language-based understanding. Deploying them on monocular video raises two
problems. Pose detection yields 2D joint sequences, leaving a dimensional gap
from the 3D motion used in training. Moreover, the 2D observation changes with
the camera viewpoint, while the depth lost in projection makes the 2D pose
ambiguous across viewpoints. We propose MoPort, a view-aware 2D interface that
enables MoLMs to take 2D observations as input. We introduce Paired View
Alignment (PVA), which gives two views of the same motion a shared 3D teacher
target and directly constrains the consistency of their latents, so that 2D
observations from different viewpoints are mapped to a consistent semantic
representation. Camera-Modulated Encoding (CME) estimates the camera
extrinsics from the 2D sequence and uses them to modulate
encoding, providing the 2D latents with geometric information about the
viewpoint. MoPort achieves R@1 of 0.5674 on HumanML3D and 0.3286 on KIT-ML,
outperforming all evaluated baselines. On HumanML3D, its R@1
decreases only from 0.5763 on in-distribution viewpoints to 0.5585 on
out-of-distribution viewpoints. We validate portability on three
downstream MoLMs. Furthermore, on the real-world RGB video dataset Fit3D,
MoPort achieves 80.36\% accuracy. Code will be released upon acceptance.
\end{abstract}

\section{Introduction}
\label{sec:introduction}

Human motion understanding is foundational to human-centric intelligent
applications, including fitness coaching~\cite{fieraru2021aifit}, dance
analysis~\cite{li2021aistpp}, and embodied AI~\cite{fu2025humanplus}.
In recent years, this field has made substantial progress
alongside the extension of large language models to multimodal
tasks~\cite{li2023blip2,liu2023llava,jiang2023motiongpt,chen2024motionllm}.
In particular, motion language models (MoLMs), such as
MotionGPT~\cite{jiang2023motiongpt} and MotionLLM~\cite{chen2024motionllm},
connect human motion with natural language. MoLMs usually encode 3D human motion into discrete tokens and model motion-text
correspondences at the token level, supporting tasks such as motion
captioning, motion Question Answering (Q\&A), and Text-to-Motion (T2M) generation.

Existing MoLMs rely on 3D body-joint sequences collected using motion-capture (MoCap)
systems for training. This representation is compact and independent of
texture and appearance, but such systems often require users to wear markers or
sensors, limiting their practicality in everyday settings. Monocular cameras
are instead easier to deploy, but provide only 2D RGB image sequences. This
causes a data domain gap between training and deployment. One solution is to
reconstruct 3D motion from
video~\cite{kanazawa2018hmr,goel2023humans,shin2024wham,shen2024gvhmr}, or to
first detect 2D keypoints and lift them to
3D~\cite{martinez2017simple,pavllo2019videopose3d,zhu2023motionbert}. However,
these pipelines add computation cost beyond 2D detection, and
reconstruction errors can propagate into motion-semantic alignment.

Developing a 2D input interface for MoLMs provides a more practical alternative to
these 3D pipelines.
2D-Interface~\cite{yokoyama2026interface} aligns 2D observations with the latent
features of a pretrained 3D encoder, allowing a pretrained MoLM to reuse its motion
understanding capabilities. However, the 2D projection of a motion changes with
the viewpoint. 2D-Interface does not model or utilize camera viewpoint
information, and its motion understanding accuracy drops when observed
viewpoints deviate substantially from the training distribution.

To address these problems, we propose MoPort, a view-aware 2D interface for MoLMs, as shown in Figure~\ref{fig:teaser}.
MoPort maps 2D joint sequences directly into the native latent and token
spaces of a pretrained MoLM, without reconstructing 3D motion or retraining the
downstream models. Our motivation is that 2D observations of the same motion
under different viewpoints should map to a consistent semantic
representation in latent space, while camera viewpoint information can explain the
differences among observations. Accordingly, we build MoPort around two components: Paired View Alignment (PVA) and Camera-Modulated Encoding (CME).

PVA constrains the encoded representations to remain consistent across
views. We project each 3D motion into two independently sampled views and
use the latents of the pretrained 3D encoder and the native tokens as
shared targets. The two observations can retain different deviations when
fitted separately to the teacher. Therefore, we further utilize a
view loss to bring their latents closer.

Projecting 3D motion to 2D observations loses depth information, leaving 2D joints with
ambiguity~\cite{wandt2021canonpose,wandt2022elepose}.
For example, an arm extending upward in the image may be raised under a
level camera or reaching forward under a downward-looking camera. Camera
extrinsics can provide geometric context for encoding the 2D
observation. To this end, we propose CME to estimate extrinsics from the 2D
sequence, use them to modulate encoding, and transform the observation geometry
of each joint into a shared reference frame.

Our contributions are threefold:
\begin{itemize}
  \item We present MoPort, a view-aware interface that connects 2D joints to the token space of pretrained MoLMs.
  \item We introduce PVA to enforce cross-view representation consistency and
  CME to utilize viewpoint-dependent observation geometry during encoding.
  \item Extensive quantitative and qualitative experiments show that MoPort
  enables pretrained MoLMs to produce accurate and view-consistent captions from 2D
  observations, adapting to different MoLMs and outperforming baselines that
  rely on 3D estimation. We further validate MoPort on the
  real-world RGB video dataset Fit3D.
\end{itemize}

\begin{figure*}[t]
  \centering
  \includegraphics[width=\textwidth]{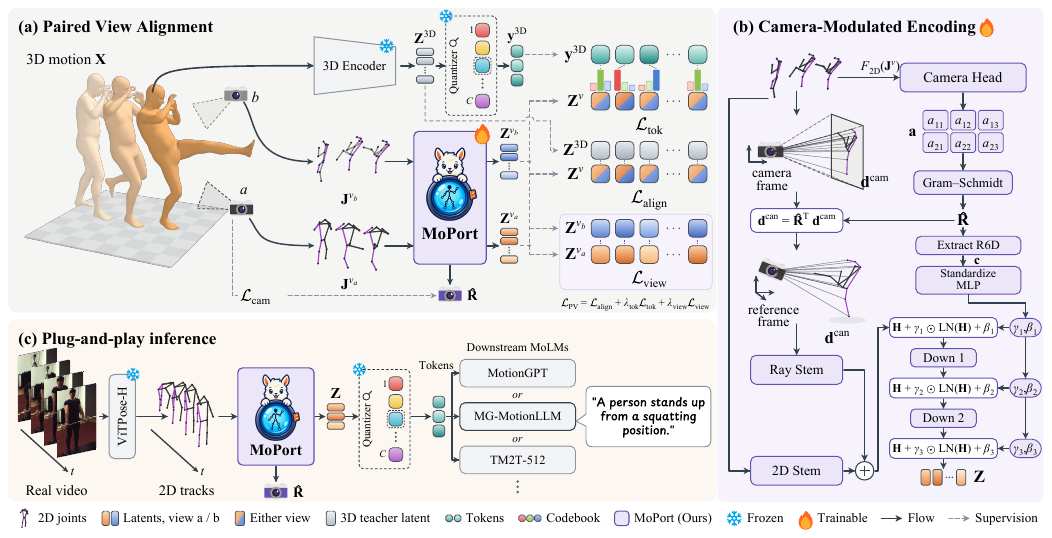}
  \caption{\textbf{MoPort framework.}
  (a) A shared MoPort encodes different 2D projections of one 3D motion and
  aligns them with latent and token targets from a pretrained 3D encoder.
  The view loss pulls the two latents together, and sampled cameras
  supervise the camera head.
  (b) The camera head estimates the extrinsics as $\mathbf{a}$, and
  Gram-Schmidt gives $\widehat{\mathbf{R}}$, which rotates the rays into the
  3D reference frame. An MLP maps the standardized R6D representation
  $\mathbf{c}$ of $\widehat{\mathbf{R}}$ to multi-scale FiLM parameters.
  (c) At inference, YOLOv8x~\cite{jocher2023yolov8} and
  ViTPose-H~\cite{xu2022vitpose} extract 2D tracks, which MoPort
  maps to the native tokens of a pretrained MoLM. Dashed arrows indicate
  training supervision.}

  \label{fig:moport_framework}
\end{figure*}

\section{Related Work}
\label{sec:related_work}

\subsection{Motion-Language Representations}
HumanML3D and KIT-ML pair MoCap sequences with natural-language
descriptions for motion-language modeling~\cite{mahmood2019amass,guo2022humanml3d,plappert2016kit}. The field includes
T2M generation~\cite{tevet2023mdm,zhang2023t2mgpt,guo2024momask}, text-motion
retrieval~\cite{petrovich2023tmr}, motion Q\&A~\cite{endo2023motionqa}, etc. Vector
quantization~\cite{vandenoord2017vqvae} compresses 3D motion into a discrete
codebook that language models can process. Building on this, TM2T casts captioning and generation as translation between token
sequences~\cite{guo2022tm2t}, MotionGPT treats quantized motion as a language
vocabulary~\cite{jiang2023motiongpt}, and MG-MotionLLM extends motion
understanding across annotation granularities~\cite{wu2025mgmotionllm}.
AvatarGPT and M$^3$GPT unify understanding, planning, and
generation~\cite{zhou2024avatargpt,luo2024m3gpt}, while MotionLLM combines
motion and video~\cite{chen2024motionllm}. FiGMo builds
an LLM-based model whose pose encoder accepts both 2D and 3D
skeletons~\cite{markhorst2026figmo}. MoPort connects to the 3D
motion input route, preserving their pretrained token space
and learning a mapping from 2D observations into it.

\subsection{Adapters Based on Representation Alignment}
Lightweight adapters align representations across modalities to make new
inputs accessible to pretrained models. BLIP-2 bridges frozen image encoders
and language models with a lightweight Querying Transformer~\cite{li2023blip2}. LLaVA trains a projection layer during feature alignment
before visual instruction tuning~\cite{liu2023llava}.
2D-Interface trains a 2D encoder and adapters against pretrained
MoLMs, aligning 2D observations with a pretrained 3D
encoder's latent space~\cite{yokoyama2026interface}. Detected 2D keypoints~\cite{xu2022vitpose} can thus bypass full 3D reconstruction.
MoPort follows this paradigm and additionally uses paired views and
estimated extrinsics~\cite{wandt2021canonpose,zhou2019continuity} to assist
cross-view latent alignment.

\subsection{Monocular 3D Motion Estimation}
2D-to-3D lifting spans per-frame regression, temporal convolutions, and
spatio-temporal transformers~\cite{martinez2017simple,pavllo2019videopose3d,zheng2021poseformer}.
MotionBERT and MotionAGFormer lift 2D sequences using learned motion
representations~\cite{zhu2023motionbert,mehraban2024motionagformer}, while
D3DP averages diffusion-generated 3D hypotheses at the pose level
(P-Agg)~\cite{shan2023d3dp}. RGB methods additionally exploit image
evidence: HMR, 4DHumans, and SAM-3D-Body estimate body meshes~\cite{kanazawa2018hmr,goel2023humans,yang2026sam3dbody}. TRACE jointly tracks
people and estimates global 3D trajectories~\cite{sun2023trace}. WHAM
recovers world-grounded motion~\cite{shin2024wham}, and GVHMR estimates
motion in gravity-aligned view coordinates~\cite{shen2024gvhmr}. Ray3D converts
2D keypoints into normalized rays and takes camera extrinsics as
input~\cite{zhan2022ray3d}, whereas CME estimates the extrinsics from the 2D
sequence.
These estimators recover 3D poses or motion. Pr-VIPE instead learns
view-invariant probabilistic embeddings of 2D poses~\cite{sun2020prvipe},
whereas MoPort learns the native motion-language representation directly.
\section{Method}
\label{sec:method}

This section presents the technical formulation of MoPort, with an overview in
Figure~\ref{fig:moport_framework}. We first formulate the problem in
Section~\ref{sec:problem}, then introduce Paired View Alignment (PVA) in
Section~\ref{sec:paired_view_alignment} and Camera-Modulated Encoding (CME) in
Section~\ref{sec:camera_conditioning}, and finally describe its portability
to different downstream MoLMs in Section~\ref{sec:integration}.

\subsection{Problem Formulation}
\label{sec:problem}

\paragraph{Setup and notation.}
Let $\mathbf{X}\in\mathbb{R}^{T\times D}$ be a length-$T$ motion feature
sequence whose $D$ channels encode root motion, local joint positions and
rotations, joint velocities, and foot contacts. A camera
with known intrinsics $\mathbf{K}$ observes the motion from view $v$, giving
the 2D observation $\mathbf{J}^{v}\in\mathbb{R}^{T\times13\times2}$ of 13
COCO joints~\cite{lin2014coco}. The feature builder $F_{2D}$ maps
$\mathbf{J}^{v}$ to 68D features encoding hip motion, relative joint
positions, planar angles, and velocities. We assume access to a pretrained
MoLM whose 3D encoder $E_{3D}$ maps $\mathbf{X}$ to the latent sequence
$\mathbf{Z}^{3D}=E_{3D}(\mathbf{X})\in\mathbb{R}^{S\times512}$, where $S$ is
the temporally downsampled length. Its native codebook $\mathbf{B}$
quantizes the latents into tokens for its language model.
\paragraph{Goal.}
We aim to learn a 2D interface
\begin{equation*}
  \mathbf{Z}^{v}=\operatorname{MoPort}_{\theta}(\mathbf{J}^{v};\mathbf{K})
  \in\mathbb{R}^{S\times512}
\end{equation*}
that replaces $E_{3D}$ at deployment, so that the pretrained MoLM understands
2D observations without modification. Because $\mathbf{B}$ and the language
model are frozen, $\mathbf{B}$ should quantize $\mathbf{Z}^{v}$ into the same
tokens as $\mathbf{Z}^{3D}$ for different views of the same motion. For
example, the stair descent in Figure~\ref{fig:table1_humanml3d_cases} should
yield the same tokens from every camera. A direct
formulation fits the teacher view by view,
\begin{equation}
  \min_{\theta}\;
  \mathbb{E}_{\mathbf{X},\,v}\,
  \ell\big(\operatorname{MoPort}_{\theta}(\mathbf{J}^{v};\mathbf{K}),
  \mathbf{Z}^{3D}\big),
  \label{eq:per_view_goal}
\end{equation}
where $\ell$ measures latent and token discrepancies.
At inference, MoPort receives only the observed 2D sequence and the known
intrinsics $\mathbf{K}$.
\subsection{Paired View Alignment}
\label{sec:paired_view_alignment}

Instead of fitting views independently as in Eq.~\eqref{eq:per_view_goal}, we
randomly sample two independent camera views $v_a$ and $v_b$ for every
training motion. Both projections share the same 3D latent and token targets.
Therefore, we align each observation with the pretrained 3D teacher,
\begin{equation}
  \mathcal{L}_{\mathrm{align}}=
  \frac{1}{2}\sum_{v\in\{v_a,v_b\}}
  \left\|\mathbf{Z}^{v}-\mathbf{Z}^{3D}\right\|_1,
  \label{eq:align_loss}
\end{equation}
and retain the native codebook supervision,
\begin{equation}
  \mathcal{L}_{\mathrm{tok}}=-\frac{1}{2}
  \sum_{v\in\{v_a,v_b\}}
  \log p(\mathbf{y}^{3D}\mid\mathbf{Z}^{v},\mathbf{B}),
  \label{eq:tok_loss}
\end{equation}
where $\mathbf{y}^{3D}$ denotes the token sequence assigned by the pretrained native
codebook. The distribution $p$ is the softmax of negative squared distances to codebook
entries, divided by the training temperature $\tau_{\mathrm{train}}$. 

Furthermore, to reduce the ambiguity introduced by different projection viewpoints, we directly pull the paired latent representations together,
\begin{equation}
  \mathcal{L}_{\mathrm{view}}=
  \left\|\mathbf{Z}^{v_a}-\mathbf{Z}^{v_b}\right\|_1.
  \label{eq:view_loss}
\end{equation}
The PVA objective is
\begin{equation}
  \mathcal{L}_{\mathrm{PV}}=\mathcal{L}_{\mathrm{align}}
  +\lambda_{\mathrm{tok}}\mathcal{L}_{\mathrm{tok}}
  +\lambda_{\mathrm{view}}\mathcal{L}_{\mathrm{view}},
  \label{eq:pv_objective}
\end{equation}
with $\lambda_{\mathrm{tok}}=0.1$ and $\lambda_{\mathrm{view}}=0.5$.

\begin{table*}[t]
  \centering
  \caption{\textbf{Comparison on HumanML3D.} All routes use pretrained
  MG-MotionLLM. Overall averages ID and OOD views. - marks a non-applicable
  cell. In all tables, bold marks the best per-column value among
  non-privileged methods, per dataset.}
  \label{tab:so3_main}
  \scriptsize
  \setlength{\tabcolsep}{1.4pt}
  \resizebox{\textwidth}{!}{%
  \begin{tabular}{llccccccccccccccc}
    \toprule
    \multirow{2}{*}{Method} & \multirow{2}{*}{Input} &
      \multicolumn{6}{c}{Overall} & \multicolumn{4}{c}{OOD} &
      \multicolumn{4}{c}{ID} &
      \multirow{2}{*}{\shortstack{PA-MPJPE\\(mm) $\downarrow$}} \\
    \cmidrule(lr){3-8}\cmidrule(lr){9-12}\cmidrule(lr){13-16}
     & & R@1 $\uparrow$ & R@2 $\uparrow$ & R@3 $\uparrow$ & MM-D $\downarrow$
       & BERTScore $\uparrow$ & CIDEr $\uparrow$
       & R@1 $\uparrow$ & R@2 $\uparrow$ & R@3 $\uparrow$ & MM-D $\downarrow$
       & R@1 $\uparrow$ & R@2 $\uparrow$ & R@3 $\uparrow$ & MM-D $\downarrow$ & \\
    \midrule
    GT3D & Privileged & 0.5924 & 0.7812 & 0.8643 & 2.5513
      & 0.4227 & 0.6457
      & - & - & - & - & - & - & - & - & - \\
    \midrule
    MotionBERT~\cite{zhu2023motionbert} & 2D$\rightarrow$3D
      & 0.0845 & 0.1499 & 0.2056 & 7.8712
      & 0.1867 & 0.0738
      & 0.0705 & 0.1267 & 0.1762 & 8.3220
      & 0.0985 & 0.1732 & 0.2351 & 7.4203 & 162 \\
    MotionAGFormer~\cite{mehraban2024motionagformer} & 2D$\rightarrow$3D
      & 0.0529 & 0.0981 & 0.1399 & 8.7388
      & 0.1636 & 0.0540
      & 0.0448 & 0.0857 & 0.1256 & 9.0162
      & 0.0609 & 0.1105 & 0.1542 & 8.4615 & 194 \\
    D3DP-P-Agg~\cite{shan2023d3dp} & 2D$\rightarrow$3D
      & 0.0343 & 0.0657 & 0.0972 & 9.6378
      & 0.1585 & 0.0455
      & 0.0330 & 0.0639 & 0.0962 & 9.6884
      & 0.0356 & 0.0674 & 0.0981 & 9.5872 & 171 \\
    \midrule
    WHAM~\cite{shin2024wham} & RGB$\rightarrow$3D
      & 0.2559 & 0.3844 & 0.4696 & 5.4994
      & 0.2698 & 0.2305
      & 0.1936 & 0.3007 & 0.3780 & 6.3281
      & 0.3182 & 0.4682 & 0.5611 & 4.6707 & \best{93} \\
    GVHMR~\cite{shen2024gvhmr} & RGB$\rightarrow$3D
      & 0.2419 & 0.3620 & 0.4429 & 5.8737
      & 0.2586 & 0.2216
      & 0.2009 & 0.3052 & 0.3803 & 6.4700
      & 0.2829 & 0.4187 & 0.5055 & 5.2774 & 99 \\
    TRACE~\cite{sun2023trace} & RGB$\rightarrow$3D
      & 0.0816 & 0.1379 & 0.1864 & 8.8676
      & 0.1863 & 0.0862
      & 0.0683 & 0.1200 & 0.1669 & 8.9482
      & 0.0950 & 0.1558 & 0.2059 & 8.7870 & 118 \\
    SAM-3D-Body~\cite{yang2026sam3dbody} & RGB$\rightarrow$3D
      & 0.0878 & 0.1470 & 0.1972 & 8.5260
      & 0.1867 & 0.0845
      & 0.0713 & 0.1255 & 0.1742 & 8.7187
      & 0.1043 & 0.1685 & 0.2203 & 8.3334 & 103 \\
    \midrule
    2D-Interface~\cite{yokoyama2026interface} & 2D joints
      & 0.4537 & 0.6218 & 0.7071 & 3.7025
      & 0.3152 & 0.3673
      & 0.3957 & 0.5491 & 0.6314 & 4.2938
      & 0.5118 & 0.6944 & 0.7828 & 3.1112 & - \\
    \textbf{MoPort (Ours)} & 2D joints
      & \best{0.5674} & \best{0.7525} & \best{0.8378} & \best{2.7274}
      & \best{0.4091} & \best{0.6119}
      & \best{0.5585} & \best{0.7438} & \best{0.8296} & \best{2.7878}
      & \best{0.5763} & \best{0.7612} & \best{0.8460} & \best{2.6670} & - \\
    \bottomrule
  \end{tabular}%
  }
\end{table*}

\subsection{Camera-Modulated Encoding}
\label{sec:camera_conditioning}

PVA constrains output consistency across views, but it does not provide
explicit observation-viewpoint information during encoding. To address this,
we introduce CME to supply observation geometry during encoding. Given known
intrinsics $\mathbf{K}$, the camera head estimates the extrinsics, which we
parameterize as $\mathbf{a}\in\mathbb{R}^{6}$ following the continuous 6D
rotation representation (R6D) of Zhou et al.~\cite{zhou2019continuity}. This
parameterization lets the head regress freely in $\mathbb{R}^{6}$ while
always yielding an orthonormal matrix. The camera head
$h_{\mathrm{cam}}$ receives motion features and outputs $\mathbf{a}$.
Gram-Schmidt orthonormalization then gives
$\widehat{\mathbf{R}}=\operatorname{GS}(\mathbf{a})$, the extrinsics that map
the 3D reference frame used to generate training projections to the camera
frame.
Sampled cameras supervise $\widehat{\mathbf{R}}$ during training. In both training and
inference, the predicted extrinsics drive the ray transformation in
Eq.~\eqref{eq:canonical_ray} and the encoder modulation in
Eq.~\eqref{eq:camera_film}, where the latter uses the R6D representation
$\mathbf{c}\in\mathbb{R}^{6}$ of $\widehat{\mathbf{R}}$ as the condition.
The R6D condition is standardized with the training-set mean and standard
deviation before entering the modulation MLP.

We complement the global extrinsics with joint-wise viewing geometry.
Coordinates $(u_{tj},v_{tj})$ locate joint $j$ at frame $t$ on a unit-focal
image plane centered at the principal point, with $u$ rightward and $v$
downward. For pixel coordinates $(x,y)$, preprocessing uses
$(u,v)=((x-c_x)/f_x,(y-c_y)/f_y)$, where $(f_x,f_y)$ and $(c_x,c_y)$
are focal lengths and principal point. The ray branch uses these observation coordinates before the
68D motion-feature construction, and rotates their directions into
the shared 3D reference frame,
\begin{equation}
  \mathbf{d}^{\mathrm{cam}}_{tj}=
  \frac{[u_{tj},-v_{tj},1]^{\top}}
       {\|[u_{tj},-v_{tj},1]\|_2},\qquad
  \mathbf{d}^{\mathrm{can}}_{tj}=
  \widehat{\mathbf{R}}^{\top}\mathbf{d}^{\mathrm{cam}}_{tj}.
  \label{eq:canonical_ray}
\end{equation}

At each frame, the ray feature concatenates the 13 canonical directions, their
offsets from the midpoint of the two hip rays, and their temporal differences,
giving $117$ dimensions. These directions remain
view-dependent, expressed in shared axes. A temporal input mapping, the ray
stem, projects the standardized ray features to the hidden width and adds
them to the 2D motion-feature stem.

The R6D condition controls how this geometric content is processed. An MLP predicts
scale and bias at the input stem and at both temporal downsampling blocks. Feature-wise linear modulation (FiLM)~\cite{perez2018film} uses these
camera-dependent parameters
to scale and shift hidden features. For hidden state $\mathbf{H}_{i}$,
the residual update is
\begin{equation}
  \widetilde{\mathbf{H}}_{i}=\mathbf{H}_{i}
  +\boldsymbol{\gamma}_{i}(\mathbf{c})\odot
   \operatorname{LN}(\mathbf{H}_{i})
  +\boldsymbol{\beta}_{i}(\mathbf{c}).
  \label{eq:camera_film}
\end{equation}
Here $\operatorname{LN}$ is layer normalization and $\odot$ denotes
element-wise multiplication.
The ray stem and FiLM output layers are zero-initialized, so the ray and
modulation branches start as identity residuals. R6D represents the
extrinsics, while canonical rays describe how they act on each
observed joint.

The camera loss $\mathcal{L}_{\mathrm{cam}}$ is the geodesic distance between
the predicted and sampled-camera extrinsics. The complete objective adds it
to Eq.~\eqref{eq:pv_objective},
\begin{equation}
  \mathcal{L}=\mathcal{L}_{\mathrm{PV}}+\lambda_{\mathrm{cam}}\mathcal{L}_{\mathrm{cam}},
  \label{eq:full_objective}
\end{equation}
with $\lambda_{\mathrm{cam}}=0.1$. The camera head, ray stem, FiLM layers, and 2D encoder are optimized jointly.
The 3D teacher and all downstream model components remain frozen.

\subsection{Plug-and-Play Portability to Downstream MoLMs}
\label{sec:integration}

MoPort replaces only the 3D encoder of a pretrained MoLM. For a given MoLM,
we train one interface against its native pretrained 3D encoder and codebook
with the same architecture and training procedure, and MoLMs that share a
tokenizer can share one interface. At deployment, MoPort encodes the 2D
sequence, and the native codebook quantizes its latents into
input tokens for the downstream MoLM, whose language model stays unchanged.
\section{Experiments}
\label{sec:experiments}
\begin{figure*}[t]
  \centering
  \includegraphics[width=\textwidth]{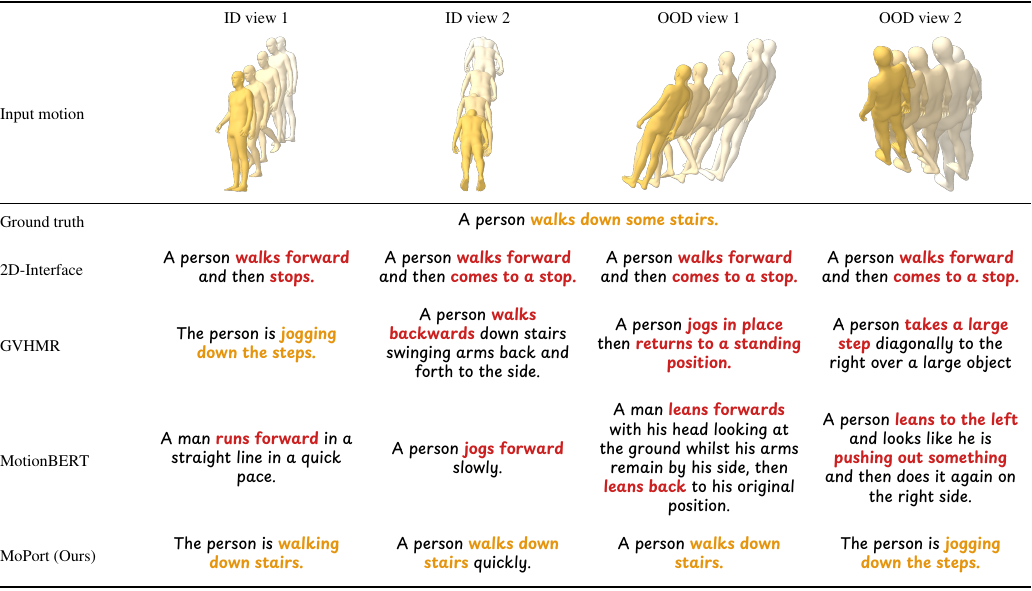}
  \caption{\textbf{MoPort describes the stair descent from all four views.}
  Other routes miss the descent or change with the camera. Captions use
  native variable-length sequence processing.
  Each column corresponds to one of two ID or two OOD cameras. The top row
  shows the motion in temporal order from light to dark.
  MoPort and 2D-Interface receive 2D joints, GVHMR reconstructs 3D motion
  from RGB video, and MotionBERT lifts 2D joints to 3D; all routes use
  MG-MotionLLM.
  Amber marks correct stair-descent descriptions, and red marks incorrect actions.}
  \label{fig:table1_humanml3d_cases}
\end{figure*}

\subsection{Setup}
\label{sec:experiment_setup}

\paragraph{Implementation details.}
Training views follow the camera distribution of
2D-Interface~\cite{yokoyama2026interface}: yaw about the vertical axis is
drawn uniformly from integer angles in $[-180^\circ,179^\circ]$, pitch from
integer angles in $[0^\circ,59^\circ]$, and roll is fixed at $0^\circ$.
We use the native variable-length sequence processing and temporal
downsampling of each benchmark and downstream MoLM~\cite{guo2022humanml3d}.
For trained interfaces, Tables~\ref{tab:so3_main}--\ref{tab:component_matrix}
report MoPort means over three training seeds. For the camera-representation
comparison, Table~\ref{tab:camera_representation} reports the mean and
standard deviation over 20 runs with resampled test cameras to show the
stability of the gains.

\paragraph{Baselines.}
We use pretrained MG-MotionLLM as the default MoLM.
2D-Interface uses the officially released
encoder~\cite{yokoyama2026interface}. MotionBERT, MotionAGFormer, and D3DP-P-Agg
receive the same 2D projections as MoPort with the preprocessing in their
official implementations~\cite{zhu2023motionbert,mehraban2024motionagformer,shan2023d3dp}.
WHAM, GVHMR, TRACE, and SAM-3D-Body receive 512$\times$512 videos under the same
cameras~\cite{shin2024wham,shen2024gvhmr,sun2023trace,yang2026sam3dbody}.
All recovered 3D skeletons are mapped to HumanML3D features. GT3D provides a reference
with ground-truth 3D input to the same downstream model.

\paragraph{Datasets.}   
We evaluate on two motion-language benchmarks, HumanML3D~\cite{guo2022humanml3d}
and KIT-ML~\cite{plappert2016kit}. Both use the official splits, including
mirrored copies: HumanML3D has 23,384, 1,460, and 4,648 motions for training,
validation, and testing, and KIT-ML has 4,888, 300, and 830. MoPort is
trained on each training split.
         
In-distribution (ID) observations use the training distribution. Out-of-distribution
(OOD) observations use
camera rotations randomly sampled from SO(3) outside the ID range. More specifically, yaw is
sampled uniformly over $[-180^\circ,180^\circ)$, while pitch leaves the ID range for
$[-60^\circ,-10^\circ]$ or $[70^\circ,85^\circ]$, roll is sampled within
$\pm[15^\circ,45^\circ]$. Each test motion yields one
single-view observation from each of two ID and two OOD cameras, and every
observation is encoded independently.

\paragraph{Evaluation metrics.}
R@k and MM-Dist measure caption-motion matching with the text-motion
matching evaluator that Guo et al.~\cite{guo2022humanml3d} pretrained for
each dataset, independently of the MoLM that generates the caption. Within each group of 32 motions, we rank
ground-truth motion embeddings by Euclidean distance from the generated
caption embedding. R@k is the fraction of captions whose ground-truth motion ranks
in the top $k$, so chance R@1 is $1/32\approx0.031$. MM-Dist, abbreviated as MM-D in tables, is the mean distance between each generated caption
and its ground-truth motion. BERTScore F1~\cite{zhang2020bertscore} and
CIDEr~\cite{vedantam2015cider} use the original human captions. Overall
averages the four views equally, while GT3D is scored once per motion.

As a reference, we additionally report the Procrustes-aligned mean per-joint
position error (PA-MPJPE) for the 3D routes. MoPort and 2D-Interface do not
reconstruct 3D joints, so the metric does not apply to them. 

\subsection{Main Results}
\label{sec:main_results}

\paragraph{Comparisons on HumanML3D.}

Table~\ref{tab:so3_main} compares the input routes on HumanML3D. MoPort
reaches Overall R@1 of 0.5674 and MM-Dist of 2.7274, recovering 95.8\% of
the GT3D R@1 of 0.5924. It leads 2D-Interface by 0.1137 R@1 and closes 82\% of the gap between
2D-Interface and GT3D. The gain concentrates on unseen viewpoints: from ID
to OOD views, the R@1 of MoPort falls by only 3.1\%, from 0.5763 to 0.5585,
whereas that of 2D-Interface falls by 22.7\%, from 0.5118 to 0.3957.
MoPort is also higher on ID views.

The 3D routes remain far below both direct interfaces. WHAM leads the RGB
methods at 0.2559 R@1 and MotionBERT leads the lifting methods at 0.0845,
while D3DP-P-Agg reaches only 0.0343, close to the chance level of 0.031. Viewpoint
sensitivity is not specific to 2D interfaces. Every 3D route except
D3DP-P-Agg, which is already near chance, loses 26--39\% of its R@1 from ID
to OOD views, more than either direct 2D interface. Geometric accuracy does not predict
caption quality either. SAM-3D-Body and GVHMR have similar PA-MPJPE of 103
and 99 mm, yet the R@1 of GVHMR is 2.8 times higher, and D3DP-P-Agg has
a lower PA-MPJPE than MotionAGFormer, 171 versus 194 mm, but also a lower
R@1.

Figure~\ref{fig:table1_humanml3d_cases} illustrates these viewpoint effects on a stair
descent. MoPort describes walking down stairs from all four views.
2D-Interface describes walking forward and stopping from every view, which
is consistent but misses the descent, and 3D-route captions change with the
camera.

\paragraph{Comparisons on KIT-ML.}
\begin{table}[t]
  \centering
  \caption{\textbf{Supervised adaptation on KIT-ML.} All routes share one
  MG-MotionLLM fine-tuned on KIT-ML and follow native variable-length
  sequence processing.}
  \label{tab:kitml_transfer}
  \scriptsize
  \setlength{\tabcolsep}{2pt}
  \resizebox{\columnwidth}{!}{%
  \begin{tabular}{llcccccc}
    \toprule
    Method & Input & R@1 $\uparrow$ & R@2 $\uparrow$ & R@3 $\uparrow$ & MM-D $\downarrow$ & BERTScore $\uparrow$ & CIDEr $\uparrow$ \\
    \midrule
    GT3D & Privileged
      & 0.3771 & 0.5896 & 0.7197 & 3.0579 & 0.5046 & 1.2407 \\
    \midrule
    MotionBERT~\cite{zhu2023motionbert} & 2D$\rightarrow$3D
      & 0.0892 & 0.1625 & 0.2231 & 8.5425 & 0.2644 & 0.2154 \\
    WHAM~\cite{shin2024wham} & RGB$\rightarrow$3D
      & 0.1612 & 0.2773 & 0.3641 & 7.0163 & 0.3520 & 0.4893 \\
    GVHMR~\cite{shen2024gvhmr} & RGB$\rightarrow$3D
      & 0.1524 & 0.2540 & 0.3294 & 7.7564 & 0.3118 & 0.3650 \\
    2D-Interface~\cite{yokoyama2026interface} & 2D joints
      & 0.3145 & 0.4933 & 0.6070 & 4.3737 & 0.4644 & 0.9973 \\
    \textbf{MoPort (Ours)} & 2D joints
      & \best{0.3286} & \best{0.5100} & \best{0.6257} & \best{4.1658} & \best{0.4744} & \best{1.0328} \\
    \bottomrule
  \end{tabular}%
  }
\end{table}

Table~\ref{tab:kitml_transfer} evaluates supervised adaptation of the same
interface training procedure to another motion corpus and annotation
vocabulary. MoPort improves R@1/2/3 over
2D-Interface by 0.0141/0.0167/0.0187, lowers MM-Dist by 0.2079, and also
leads on BERTScore and CIDEr.
This margin is smaller than the 0.1137 on
HumanML3D, and MoPort recovers 87.1\% of the GT3D R@1, compared with
95.8\% on HumanML3D.

\subsection{Ablation Study}
\label{sec:ablation_study}

\begin{figure}[t]
  \centering
  \includegraphics[width=\columnwidth,trim=0 0.24in 0 0,clip]{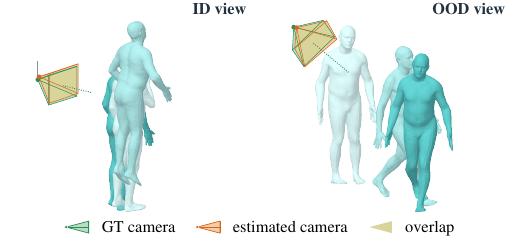}\par
  \vspace{1pt}
  \begin{minipage}[c][20pt][c]{0.49\columnwidth}
    \centering\fontsize{8.5}{10}\selectfont
    A person jumps vertically.
  \end{minipage}\hfill
  \begin{minipage}[c][20pt][c]{0.49\columnwidth}
    \centering\fontsize{8.5}{10}\selectfont
    A person walks, turns around,\\
    and walks back.
  \end{minipage}\par
  \vspace{1pt}
  \includegraphics[width=\columnwidth,trim=0 0 0 1.44in,clip]{figures/camera_posterior/moport_camera_posterior_best_cases_noerr.pdf}
  \caption{\textbf{MoPort estimates extrinsics in ID and OOD views.}
  Left: ID case. Right: OOD case. Green and
  amber frustums are the ground-truth and estimated cameras, overlapping in
  khaki. Body color darkens over time.}
  \label{fig:camera_posterior}
\end{figure}

\paragraph{Training-recipe comparison.}
\begin{table}[t]
  \centering
  \caption{\textbf{Ablation study on HumanML3D and KIT-ML.}
  Baseline results are reused from the two main comparison tables,
  Tables~\ref{tab:so3_main} and~\ref{tab:kitml_transfer}. Within each dataset,
  all variants share the MG-MotionLLM of the corresponding main table,
  pretrained for HumanML3D and fine-tuned for KIT-ML.}
  \label{tab:component_matrix}
  \scriptsize
  \setlength{\tabcolsep}{2pt}
  \resizebox{\columnwidth}{!}{%
  \begin{tabular}{lcccccc}
    \toprule
    \multirow{2}{*}{Variant} & \multicolumn{6}{c}{Overall} \\
    \cmidrule(lr){2-7}
     & R@1 $\uparrow$ & R@2 $\uparrow$ & R@3 $\uparrow$ & MM-D $\downarrow$ & BERTScore $\uparrow$ & CIDEr $\uparrow$ \\
    \midrule
    \multicolumn{7}{l}{\textit{HumanML3D}} \\
    2D-Interface (Baseline)
      & 0.4537 & 0.6218 & 0.7071 & 3.7025 & 0.3152 & 0.3673 \\
    PVA-only
      & 0.5045 & 0.7017 & 0.8085 & 3.4399 & 0.3574 & 0.5043 \\
    CME-only
      & 0.5137 & 0.7101 & 0.8061 & 3.3554 & 0.3558 & 0.5021 \\
    MoPort
      & \best{0.5674} & \best{0.7525} & \best{0.8378} & \best{2.7274} & \best{0.4091} & \best{0.6119} \\
    \midrule
    \multicolumn{7}{l}{\textit{KIT-ML}} \\
    2D-Interface (Baseline)
      & 0.3145 & 0.4933 & 0.6070 & 4.3737 & 0.4644 & 0.9973 \\
    PVA-only
      & 0.3173 & 0.5054 & 0.6095 & 4.3063 & 0.4742 & \best{1.0379} \\
    CME-only
      & 0.3185 & 0.4995 & 0.6025 & 4.3201 & 0.4715 & 1.0158 \\
    MoPort
      & \best{0.3286} & \best{0.5100} & \best{0.6257} & \best{4.1658} & \best{0.4744} & 1.0328 \\
    \bottomrule
  \end{tabular}%
  }
\end{table}

In Table~\ref{tab:component_matrix}, variants share training settings and
the token loss $\mathcal{L}_{\mathrm{tok}}$. PVA-only adds
$\mathcal{L}_{\mathrm{align}}$ and $\mathcal{L}_{\mathrm{view}}$, while CME-only adds
the camera head, canonical rays, and R6D FiLM.

Adding PVA to CME-only improves Overall R@1 by 0.0537 on
HumanML3D and 0.0101 on KIT-ML, and adding CME to PVA-only improves it by
0.0629 and 0.0113, respectively, with lower MM-Dist in both comparisons. On
HumanML3D, each component adds a comparable gain on top of the other,
and neither subsumes the other. This matches their roles: PVA requires the
outputs for different views to agree, whereas CME tells the encoder how
each view was formed. On KIT-ML, CME-only falls slightly
below the baseline on R@3, 0.6025 versus 0.6070, and PVA-only has a higher
CIDEr than MoPort, 1.0379 versus 1.0328. MoPort leads all six metrics on HumanML3D
and five of six on KIT-ML.

\paragraph{Camera representations for CME.}
\begin{table}[t]
  \centering
  \caption{\textbf{CME variants with pretrained MG-MotionLLM.}
  Mean $\pm$ std.\ over 20 runs with resampled test cameras; alignment and
  caption metrics average four views. Oracle: true extrinsics.}
  \label{tab:camera_representation}
  \scriptsize
  \setlength{\tabcolsep}{1pt}
  \begin{tabular*}{\columnwidth}{@{\extracolsep{\fill}}lccccccc@{}}
    \toprule
     & \multicolumn{2}{c}{\textit{Camera error ($^\circ$)}} &
       \multicolumn{2}{c}{\textit{Alignment}} &
       \multicolumn{3}{c}{\textit{Caption performance}} \\
    \cmidrule(lr){2-3}\cmidrule(lr){4-5}\cmidrule(lr){6-8}
    Camera rep. & ID\,$\downarrow$ & OOD\,$\downarrow$ & \shortstack{Token\\agree.\,$\uparrow$} & \shortstack{Latent\\$\ell_1$\,$\downarrow$} &
      R@1\,$\uparrow$ & MM-D\,$\downarrow$ & CIDEr\,$\uparrow$ \\
    \midrule
    16D concat. & \valstd{11.73}{0.11} & \valstd{14.37}{0.10} & \valstd{0.3615}{0.0006} & \valstd{1.5063}{0.0014} &
      \valstd{0.5660}{0.0012} & \valstd{2.7371}{0.0083} & \valstd{0.6116}{0.0029} \\
    Quaternion & \valstd{11.61}{0.12} & \valstd{13.73}{0.12} & \valstd{0.3634}{0.0005} & \valstd{1.5083}{0.0017} &
      \valstd{0.5653}{0.0016} & \valstd{2.7475}{0.0070} & \valstd{0.6088}{0.0036} \\
    R6D & \valstd{\best{11.47}}{0.12} & \valstd{\best{13.72}}{0.10} & \valstd{\best{0.3643}}{0.0006} & \valstd{\best{1.4990}}{0.0015} &
      \valstd{\best{0.5671}}{0.0019} & \valstd{\best{2.7283}}{0.0091} & \valstd{\best{0.6118}}{0.0028} \\
    \midrule
    \begin{tabular}[c]{@{}l@{}}GT-camera\\oracle\end{tabular} & 0.00 & 0.00 & \valstd{0.3692}{0.0005} & \valstd{1.4885}{0.0015} &
      \valstd{0.5683}{0.0013} & \valstd{2.7158}{0.0073} & \valstd{0.6124}{0.0026} \\
    \bottomrule
  \end{tabular*}
\end{table}

Table~\ref{tab:camera_representation} compares three predicted-camera designs
and a GT-camera oracle. The 16D concatenation appends nine rotation-matrix
entries, six trigonometric features, and a normalized rotation magnitude to
2D inputs. Quaternion and R6D share canonical rays and three FiLM
sites, differing only in the extrinsics representation. The 16D design changes both
representation and fusion. All predicted-camera variants share random
initialization, PVA, a 2,000-step budget, and sampled-camera supervision.
Since design gaps are small, we repeat inference 20 times,
each time resampling the ID and OOD test cameras shared by all variants, to
test their significance.

Among predicted-camera designs, R6D leads every metric on HumanML3D. Its
alignment gains are significant: R6D has higher token agreement and lower
latent $\ell_1$ than both alternatives in all 20 runs, so a two-sided sign
test gives $p<10^{-5}$. Its R@1 gains over 16D and Quaternion, 0.0011 and
0.0018, are comparable to the R@1 standard deviation of 0.0012--0.0019.
Because all variants share the cameras of each run, these deviations mainly
reflect camera sampling and are similar across rows, so we compare designs
run by run.
Alignment metrics measure token agreement and latent error over four
views per motion, and camera error is the SO(3) geodesic distance to ground
truth. Camera errors rise by only $2.1^\circ$--$2.6^\circ$, from
$11.47^\circ$--$11.73^\circ$ in ID views to $13.72^\circ$--$14.37^\circ$
in OOD views.
Ground-truth extrinsics barely improve captions: the GT-camera oracle raises
token agreement from 0.3643 to 0.3692 but R@1 by only 0.0012, within one
standard deviation. Camera estimation is therefore not the main bottleneck,
and the remaining gap to GT3D more likely reflects information that 2D
joints lack, such as depth. Figure~\ref{fig:camera_posterior} shows
estimated cameras in ID and OOD cases.

\begin{figure}[t]
  \centering
  \includegraphics[width=\columnwidth]{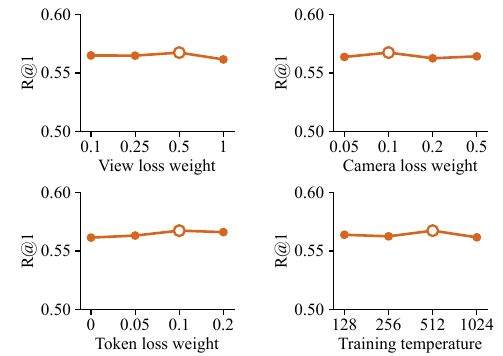}
  \caption{\textbf{R@1 varies by less than one percentage point across
  hyperparameters.} Each panel varies one setting of MoPort with
  pretrained MG-MotionLLM. Open circles mark the default.}
  \label{fig:learned_camera_hparams}
\end{figure}

\paragraph{Hyperparameter sensitivity.}
Figure~\ref{fig:learned_camera_hparams} varies one loss weight or training
temperature at a time on the same test set as
Table~\ref{tab:so3_main}. Runs share a 2,000-step budget and seed, with
defaults fixed before testing. R@1 stays within one percentage point across
all tested values, so MoPort does not depend on careful loss balancing or
temperature tuning.

\subsection{Generalizability}
\label{sec:generalizability}

\paragraph{Portability to different downstream MoLMs.}
\label{sec:portability}
\begin{table}[t]
  \centering
  \caption{\textbf{Portability across pretrained downstream MoLMs on HumanML3D.} Metrics average two ID and two
  OOD views; GT3D is scored once per motion.}
  \label{tab:three_consumers}
  \scriptsize
  \setlength{\tabcolsep}{1.5pt}
  \resizebox{\columnwidth}{!}{%
  \begin{tabular}{llcccccc}
    \toprule
    Pretrained MoLM & Input & R@1 $\uparrow$ & R@2 $\uparrow$ & R@3 $\uparrow$ & MM-D $\downarrow$ & BERTScore $\uparrow$ & CIDEr $\uparrow$ \\
    \midrule
    \multirow{2}{*}{MotionGPT~\cite{jiang2023motiongpt}} & GT3D & 0.5229 & 0.7096 & 0.7995 & 2.9831 & 0.3396 & 0.4167 \\
    & MoPort & 0.5129 & 0.6964 & 0.7859 & 3.0607 & 0.3354 & 0.4035 \\
    \midrule
    \multirow{2}{*}{TM2T-512~\cite{guo2022tm2t}} & GT3D & 0.5279 & 0.7244 & 0.8204 & 2.8880 & 0.3394 & 0.2319 \\
    & MoPort & 0.5120 & 0.7049 & 0.8011 & 3.0331 & 0.3330 & 0.2282 \\
    \midrule
    \multirow{2}{*}{MG-MotionLLM~\cite{wu2025mgmotionllm}} & GT3D & 0.5924 & 0.7812 & 0.8643 & 2.5513 & 0.4227 & 0.6457 \\
    & MoPort & \best{0.5674} & \best{0.7525} & \best{0.8378} & \best{2.7274} & \best{0.4091} & \best{0.6119} \\
    \bottomrule
  \end{tabular}%
  }
\end{table}

We evaluate portability on different downstream MoLMs, including MotionGPT~\cite{jiang2023motiongpt},
TM2T-512~\cite{guo2022tm2t}, and MG-MotionLLM~\cite{wu2025mgmotionllm}.
In Table~\ref{tab:three_consumers}, MoPort retains 98.1\%, 97.0\%, and
95.8\% of the GT3D R@1 for MotionGPT, TM2T-512, and MG-MotionLLM.
Following the setup of
2D-Interface~\cite{yokoyama2026interface}, TM2T-512 is a TM2T captioner
trained on MotionGPT's 512-entry motion tokens, sharing MotionGPT's
3D encoder and codebook. Since MoPort training doesn't involve the
language model, TM2T-512 reuses the MotionGPT interface without retraining
and still keeps 97.0\%. This indicates that MoPort
tokens remain faithful in the native token space rather than fitting one
language model.

\paragraph{Real-video adaptation on Fit3D.}
\label{sec:fit3d_real_adaptation}
\begin{table}[t]
  \centering
  \caption{\textbf{Real-video adaptation on Fit3D.}
  All routes share the same fine-tuned MG-MotionLLM. EM denotes exact match.}
  \label{tab:fit3d_real_adaptation}
  \small
  \setlength{\tabcolsep}{5pt}
  \begin{tabular}{lcc}
    \toprule
    Input route & Class accuracy (\%) $\uparrow$ & Template EM (\%) $\uparrow$ \\
    \midrule
    GT3D & 78.57 & 75.00 \\
    \midrule
    2D-Interface & 71.43 & 69.64 \\
    MoPort & \textbf{80.36} & \textbf{78.57} \\
    \bottomrule
  \end{tabular}
\end{table}
We split Fit3D~\cite{fieraru2021aifit} by subject into 560, 112, and 224
videos for training, validation, and testing, with four camera views per
recording. After fine-tuning MG-MotionLLM on 3D tokens and class names, we
freeze it together with the 3D tokenizer and then fine-tune both interfaces.

In Table~\ref{tab:fit3d_real_adaptation}, class accuracy selects one of 28
classes by length-normalized candidate likelihood, and Template EM measures
exact match between the greedy output and the class template. MoPort classifies 80.36\% of videos correctly versus 71.43\%
for 2D-Interface, an 8.93-point gain. MoPort is also 1.79 points higher
than GT3D, which reaches 78.57\%. Template EM gives the same ordering. Because the test views are real
cameras rather than sampled ones, this gain suggests that the view-aware
design carries over to real viewpoints after adaptation.
\section{Conclusion}
\label{sec:conclusion}

We have presented MoPort, a view-aware 2D interface that lets pretrained
MoLMs understand monocular 2D observations without
modification. Paired
View Alignment gives different views of the same motion a shared 3D teacher
and keeps their latents consistent, and Camera-Modulated Encoding estimates
the camera extrinsics from the 2D sequence and uses them to modulate encoding.
Together, they map different views of a motion to a consistent
semantic representation in the native token space. Experiments on
HumanML3D, KIT-ML, three downstream MoLMs, and real Fit3D videos validate
MoPort: it outperforms all evaluated baselines and remains stable on unseen
viewpoints. The GT-camera oracle further shows that camera estimation is not
the main bottleneck, and the remaining gap to GT3D points to depth cues that
2D joints lack.

\paragraph{Limitations and broader impact.} This work is limited by the
assumption of known camera intrinsics and by training mainly on synthetic
projections of MoCap data, and MoPort still fails on some
OOD views and on motions such as going down to the ground.
Future work could estimate the
intrinsics jointly, recover depth cues missing from 2D joints, and train on
large-scale real videos. MoPort could let
existing MoLMs run on ordinary cameras, for example as
fitness or rehabilitation assistants without MoCap
equipment. A potential risk is that motion understanding from
ordinary video could be misused to monitor or profile people without their
consent.
\clearpage
\section*{LLM Usage Statement}
The authors led the development of the core ideas, method design,
experiments, figure preparation, and manuscript writing, and retain
responsibility for the research decisions and final content. GPT-6-Astra
assisted with language polishing, literature search, code editing, baseline experiment
execution, and figure refinement. GPT-Image 2 was used to create the MoPort
logo. The authors take full responsibility for
the scientific claims, the accuracy of the results and citations, and
the reproducibility of the work.

{\small
\bibliographystyle{ieeenat_fullname}
\bibliography{main}
}

\appendix
\section{Supplementary Experiments}
\label{sec:supplementary_experiments}
This appendix presents implementation and training details, the Fit3D
real-video adaptation and caption-metric protocols, runtime, additional
MoPort diagnostics, and additional qualitative results.

\subsection{Implementation and Training Details}
\label{sec:implementation_details}

\paragraph{Sequence processing and pretrained models.}
Each benchmark follows the corresponding dataset and downstream MoLM's native
variable-length sequence processing, including length constraints, padding,
and temporal downsampling. The MG-MotionLLM motion encoder has two temporal
downsampling blocks of stride two, giving a total factor of four and
512D latents. The portability experiment uses each downstream
model's corresponding encoder, codebook, and language model.

\paragraph{2D interface training.}
On HumanML3D, 2D-Interface uses the authors' released MG-MotionLLM 2D encoder,
while MoPort is trained from scratch. On KIT-ML,
both 2D interfaces are trained on the same training split and camera-angle
distribution. On both datasets, interface training, camera-head training,
and component comparisons use 2D-Interface's camera
distribution~\cite{yokoyama2026interface}: yaw and pitch are drawn
uniformly from integer angles in $[-180^\circ,179^\circ]$ and
$[0^\circ,59^\circ]$, respectively, with zero roll. Intrinsics are assumed
known, and the camera head estimates extrinsics.

\paragraph{Camera head and modulation modules.}
The camera head maps each 68D frame to 256 channels with a linear layer,
layer normalization, and GELU. Three residual temporal blocks use group
normalization, depthwise convolutions of kernel size five and dilation
rates 1, 2, and 4, followed by pointwise convolutions. Temporal mean,
standard deviation, and maximum pooling are concatenated. A 768-to-256
projection, layer normalization, GELU, and a 6D output layer
predict the extrinsics. The camera head is randomly initialized and trained
jointly with the interface under the same training-camera distribution.
A kernel-three
ray convolution maps 117D features to the encoder's 512 channels. The
modulation MLP has two hidden layers of width 128, and each FiLM output
layer predicts 512D scale and bias vectors. The ray stem and
FiLM output layers are zero-initialized.

\paragraph{Optimization.}
MoPort and its variants use AdamW with betas $(0.9,0.99)$ and weight decay $10^{-4}$.
Initial learning rates are $10^{-4}$ for the 2D encoder and ray/FiLM
modules, and $5\times10^{-5}$ for the camera head, with cosine decay to
$10^{-6}$. Training uses a batch size of 64, 2,000 steps, and gradient-norm
clipping at 1. The training temperature is 512. Default
weights for the alignment, token, view, and camera losses in
Eqs.~\eqref{eq:pv_objective} and~\eqref{eq:full_objective} are
1, 0.1, 0.5, and 0.1, respectively.

\paragraph{GT-camera oracle.}
The oracle in Table~\ref{tab:camera_representation} separately trains an
R6D encoder with ground-truth extrinsics at both training
and test time, retaining 2D joints as motion input. It uses pretrained
MG-MotionLLM, PVA, and a 2,000-step budget. Evaluation uses the same
4,648 motions and four views.
Recomputed interface diagnostics verify identical tokens for all 18,592
archived observations. CIDEr uses the saved captions and the same
human-reference protocol.

\paragraph{KIT caption adapter.}
We freeze the original MG-MotionLLM weights and train only LoRA~\cite{hu2022lora} updates
to attention query and value projections, with rank 8, alpha 16, and
dropout 0.05. AdamW uses learning rate $5\times10^{-4}$, weight decay 0.01,
and batch size 32. The adapter is then frozen and shared by
GT3D, the 2D interfaces, and the 3D routes.
\subsection{Fit3D Real-Video Adaptation Protocol}
\label{sec:fit3d_real_protocol}

\paragraph{Split and supervision.}
We split the official Fit3D training partition, which provides public
3D ground truth, by subject into 560, 112, and 224 videos for training,
validation, and testing: s03, s04, s05, s07, and s08 for training, s09
for validation, and s10 and s11 for testing. We retain 28 exercise classes
with semantic names. This is a held-out-subject evaluation within the
public training partition. Supervision uses official category names in
the template \texttt{a person performs <class name>.}, replacing underscores
with spaces. These targets provide category-level action names.

\paragraph{Shared downstream model.}
We first fine-tune the language adapter of MG-MotionLLM using native tokens
from ground-truth 3D motions. The 3D encoder, codebook, normalization statistics, base language
model, and input embeddings remain frozen. Only query/value attention
LoRA parameters are updated, with rank 8, alpha 16, and dropout 0.05.
AdamW uses learning rate $5\times10^{-4}$, weight decay 0.01, global batch
size 20, and seed 274330. The budget is 50 epochs. This adapter is then
frozen and shared by all three input routes.

\paragraph{Real 2D observations.}
YOLOv8x and ViTPose-H extract 2D joints from real camera videos. We select each frame's detection by bounding-box area times
confidence and interpolate missing detections. The four cameras and 3D
ground truth are synchronized by common source-frame indices and resampled
to 20 fps. Thirteen joints form 68D features. Complete recordings retain
the native temporal stride of four, with final-frame repetition to meet
native length requirements. Dataset focal lengths and principal points
normalize the observed rays.

\paragraph{Interface fine-tuning.}
Both interfaces start from their respective HumanML3D checkpoints. For each
training recording in each epoch, we sample two of the four real cameras
uniformly without replacement. Both interfaces share the sample order and
view pairs and train for 50 epochs with global batch size 20. AdamW uses betas $(0.9,0.99)$, weight decay
$10^{-4}$, and gradient clipping at 1. Learning rates are $10^{-4}$ for
the encoder and modulation modules and $5\times10^{-5}$ for the camera head.
The 3D teacher, codebook, and fine-tuned downstream model remain frozen.

\paragraph{Evaluation.}
Both 2D interfaces use the same detections and frame indices, and GT3D uses the
corresponding 3D motion. For 28-class recognition, we score each class
template by its mean per-token negative log likelihood, including the
end token and excluding padding, and select the lowest-scoring class.
Class accuracy is the fraction of test videos classified correctly.

\begin{figure*}[t]
  \centering
  \includegraphics[width=\textwidth]{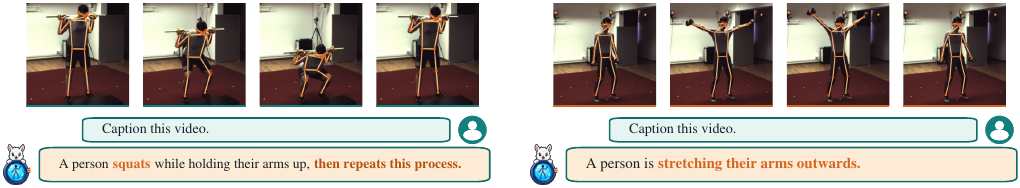}
  \caption{\textbf{MoPort captions of two Fit3D repetitions:} a squat on the left and a lateral raise on the right. Amber marks
  correct actions. Dark amber marks an incorrectly added repetition.
  Skeletons are the ViTPose-H input trajectories.}
  \label{fig:supp_fit3d_cases}
\end{figure*}

\subsection{Caption Metric Protocol}
\label{sec:caption_metrics}

BERTScore uses RoBERTa-large layer 17 with English baseline rescaling and
IDF weighting, taking the maximum F1 across human references. We match
lemmatized HumanML3D reference strings from the retrieval loader back to
the original annotation sentences. KIT-ML uses its original captions
directly. GT3D and the other methods are scored by the same rules.

CIDEr uses COCO CIDEr-D with 1- to 4-grams, $\sigma=6$, and the raw 0--10
scale. Text is lowercased and tokenized with TreebankWordTokenizer, with
non-alphanumeric tokens discarded. Both metrics fix IDF statistics on the
complete reference corpus of each evaluation cohort, counting each motion
once. Outputs are averaged within each view and then equally across
the four views. GT3D is scored once per motion.

\subsection{Runtime Analysis}
\label{sec:supp_efficiency}

Table~\ref{tab:efficiency} reports runtime from detected 2D input to
MG-MotionLLM native tokens, alongside the Overall R@1 of
Table~\ref{tab:so3_main}. We time 100 seed-fixed HumanML3D test clips from
the first ID view, 127 frames on average, at batch one on one H200, taking the median
of three runs per clip after warm-up. All timed routes start from the same rendered
RGB video. The 2D and lifting routes share 2D detection with YOLOv8x and
ViTPose-H, which takes 6.91 s per clip on average, about 54 ms per frame
including video decoding, and accounts for over 93\% of their end-to-end
time. GVHMR runs its own batched tracking and ViTPose on the 30 fps input in
4.51 s per clip including video decoding, also excluded from the table, and additionally requires HMR2~\cite{goel2023humans}
image features. Timing excludes
model loading, rendering, offline frame-rate conversion, and language decoding.

From detected 2D joints, MoPort reaches native tokens in 3.5 ms, while the
lifting routes need 28.8--457.2 ms, of which 14.5--15.5 ms is the 263D bridge
and native tokenization. GVHMR needs 1.84 s beyond its 2D detection, mostly
for HMR2 image features. End to end, 2D detection dominates, and the 2D and
lifting routes differ by at most 6.6\%. GVHMR's batched tracking and ViTPose
take 4.51 s on 1.5 times as many frames, versus 6.91 s for the per-frame
shared detector, so its end-to-end time is lower, 6.35 s versus 6.91 s for
MoPort.

\begin{table}[t]
  \centering
  \caption{\textbf{Runtime after 2D detection.} Mean time per clip at
  batch one over 100 HumanML3D test clips, from detected 2D input to
  MG-MotionLLM native tokens, next to the Overall R@1 of
  Table~\ref{tab:so3_main}, where the 2D and lifting routes take projected
  2D joints. Interface models include 2D feature construction
  and native quantization; 3D routes add the 263D bridge and native
  tokenization. GVHMR's model includes HMR2 image features. Shared 2D
  detection, 6.91 s per clip, is excluded.}
  \label{tab:efficiency}
  \scriptsize
  \setlength{\tabcolsep}{4pt}
  \begin{tabular*}{\columnwidth}{@{\extracolsep{\fill}}lcccc@{}}
    \toprule
    Route & Model (ms) & Bridge + tok.\ (ms) & Total (ms)\,$\downarrow$ & R@1\,$\uparrow$ \\
    \midrule
    2D-Interface & \best{1.9} & - & \best{1.9} & 0.4537 \\
    MoPort & 3.5 & - & 3.5 & \best{0.5674} \\
    \midrule
    MotionBERT & 13.3 & 15.5 & 28.8 & 0.0845 \\
    MotionAGFormer & 47.4 & \best{14.5} & 61.9 & 0.0529 \\
    D3DP-P-Agg & 442.4 & 14.8 & 457.2 & 0.0343 \\
    \midrule
    GVHMR & 1814.3 & 23.7 & 1838.0 & 0.2419 \\
    \bottomrule
  \end{tabular*}
\end{table}

\subsection{Additional MoPort Diagnostics}
\label{sec:supp_hparams}

The hyperparameter defaults are $\lambda_{\mathrm{view}}=0.5$, $\lambda_{\mathrm{cam}}=0.1$,
$\lambda_{\mathrm{tok}}=0.1$, and $\tau_{\mathrm{train}}=512$.

Complementing the main-text hyperparameter analysis in
Figure~\ref{fig:learned_camera_hparams},
Table~\ref{tab:supp_inference_temperature} reports inference-temperature
diagnostics on the same 4,648 test motions and four views as
Table~\ref{tab:so3_main}, averaging the four views. It uses the
default-configuration run shared by the sweeps in
Figure~\ref{fig:learned_camera_hparams}, which is trained separately from the
R6D row of Table~\ref{tab:camera_representation}, so its alignment
values differ from that table. View agreement
measures native-token agreement across views of the same motion.

\begin{table}[b]
  \centering
  \caption{\textbf{Inference-temperature sensitivity.} Temperature changes the soft
  target NLL but not the argmax tokens, so token and view agreement stay at
  0.3755 and 0.4871.}
  \label{tab:supp_inference_temperature}
  \scriptsize
  \setlength{\tabcolsep}{5.0pt}
  \begin{tabular}{rccc}
    \toprule
    $\tau_{\mathrm{infer}}$ & ID NLL\,$\downarrow$ & OOD NLL\,$\downarrow$ & Argmax hash \\
    \midrule
    128 & 15.4270 & 18.1157 & identical \\
    256 & 8.1663 & 9.5230 & identical \\
    512 & 4.8994 & 5.5885 & identical \\
    1024 & 3.7478 & 4.0971 & identical \\
    \bottomrule
  \end{tabular}
\end{table}

Inference temperature rescales token-distance logits and preserves their
argmax, so Table~\ref{tab:supp_inference_temperature} records changes in the
soft-distribution score.

\subsection{Additional Qualitative Comparisons}
\label{sec:supp_qualitative}

\begin{figure*}[t]
  \centering
  \includegraphics[width=\textwidth]{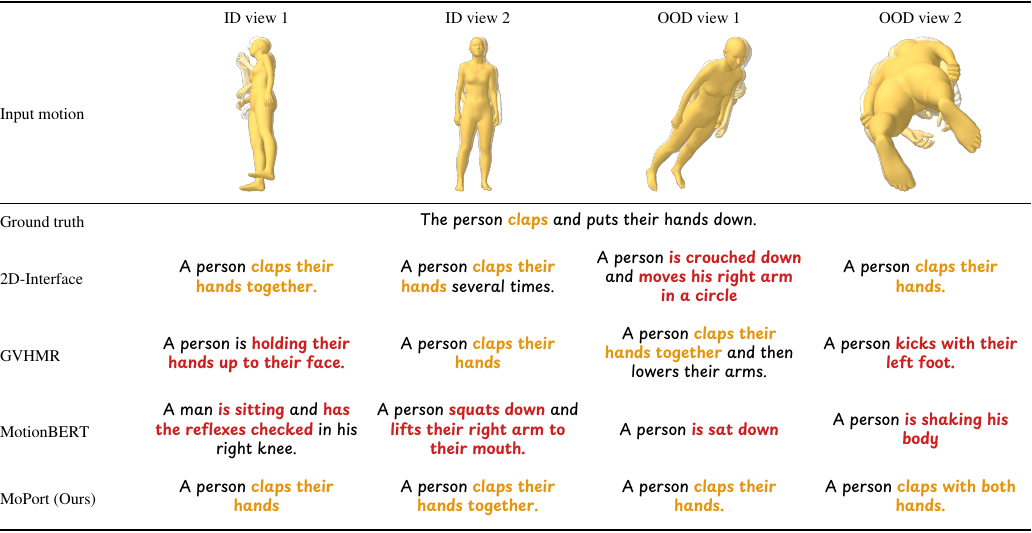}
  \caption{\textbf{Captions of a clapping motion from four viewpoints.} Layout and colors follow
  Figure~\ref{fig:table1_humanml3d_cases}.}
  \label{fig:supp_case_clap}
\end{figure*}

\begin{figure*}[t]
  \centering
  \includegraphics[width=\textwidth]{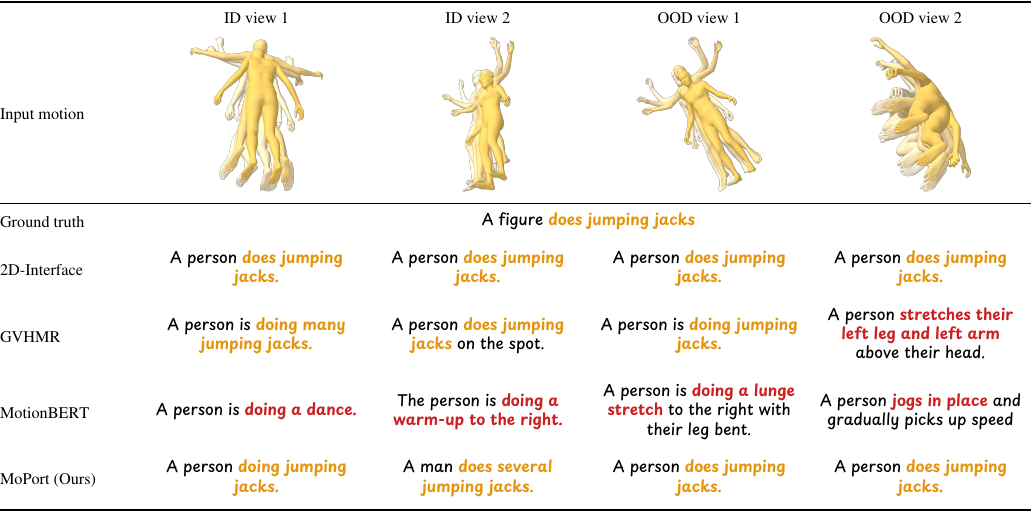}
  \caption{\textbf{Captions of a jumping-jack motion from four viewpoints.} Layout and
  colors follow Figure~\ref{fig:table1_humanml3d_cases}.}
  \label{fig:supp_case_jacks}
\end{figure*}

Figures~\ref{fig:supp_case_clap} and~\ref{fig:supp_case_jacks} repeat the layout
of Figure~\ref{fig:table1_humanml3d_cases} for two more held-out motions, with
the same routes, cameras, downstream MoLM, and correct/incorrect judgment of the main action: clapping, where MoPort is correct from all four views and 2D-Interface errs
from one, and jumping jacks, which both 2D interfaces caption correctly. Captions are the original model outputs. The qualitative cases follow the benchmark and
downstream MoLM in variable-length sequence processing and temporal
downsampling, with the corresponding human motion annotations.

\begin{figure*}[t]
  \centering
  \includegraphics[width=\textwidth]{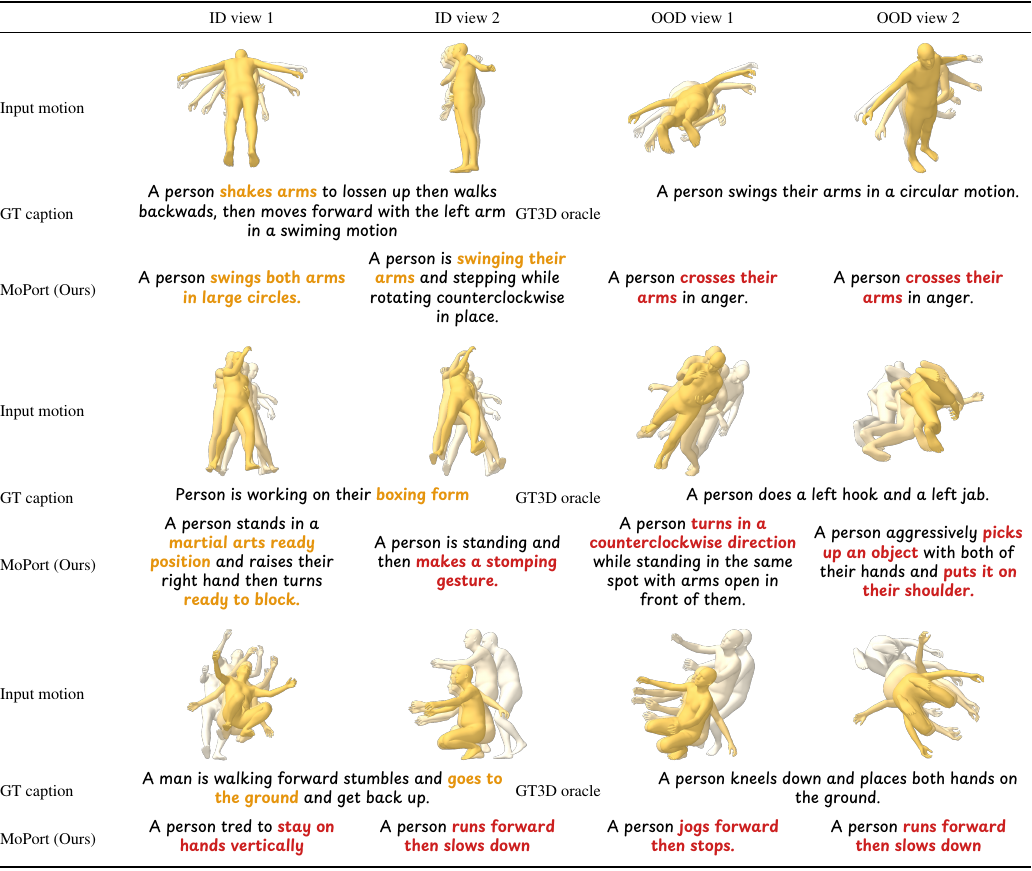}
  \caption{\textbf{Failure cases of MoPort.} Arm circles fail from the two OOD cameras, boxing is
  identified only from ID view~1, and going down to the ground is missed from
  every camera. For all three motions the pretrained MoLM captions the
  ground-truth 3D motion correctly. Each case includes the ground-truth caption, labeled GT caption, and the
  output of the pretrained MoLM from GT3D, labeled GT3D oracle. Layout and colors follow
  Figure~\ref{fig:table1_humanml3d_cases}.}
  \label{fig:supp_moport_failures}
\end{figure*}

\paragraph{Failure cases of MoPort.}
Figure~\ref{fig:supp_moport_failures} shows failed motions. The pretrained MoLM captions each correctly from ground-truth 3D
motion, while MoPort fails from at least two of the four cameras.
Arm circles are described as arm movements from the two ID cameras and as
crossed arms from both OOD cameras. Both OOD cameras lie outside
the training-camera distribution.
Boxing is identified only from ID view~1. The other ID view is
described as a stomp.
For a walk that ends on the ground, captions from all four cameras omit the
descent, and three mention only running or jogging forward.

\begin{figure*}[t]
  \centering
  \includegraphics[width=0.985\textwidth]{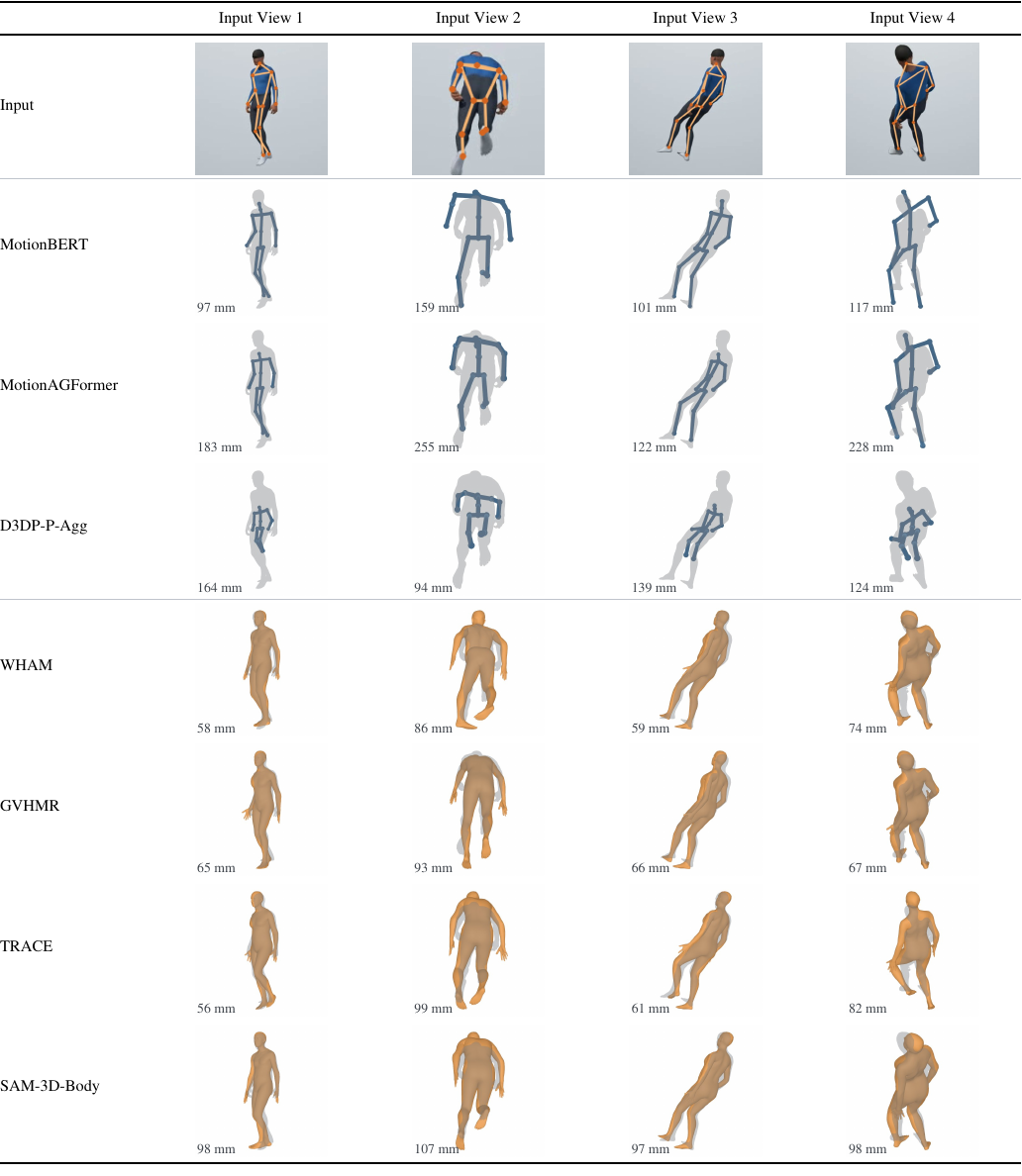}
  \caption{\textbf{3D reconstructions of one motion from four input views.}
  The top row shows rendered RGB frames and 2D joints. Blue denotes the raw
  17-joint skeletons predicted by lifting methods. Orange denotes neutral SMPL
  bodies fitted to the joints predicted by RGB methods. Translucent gray denotes
  the ground-truth body. Each column uses its corresponding input camera.
  Predictions are aligned to the ground truth in root position and global scale,
  preserving their orientation and pose. Numbers are PA-MPJPE for each view.}

  \label{fig:supp_recon_audit}
\end{figure*}

\paragraph{3D reconstructions passed to the downstream MoLM.}
Figure~\ref{fig:supp_recon_audit} shows the seven external estimators'
3D reconstructions of the motion in Figure~\ref{fig:table1_humanml3d_cases}.
The four columns correspond to its two ID and two OOD input views.
All estimators return complete sequences for all four views.
RGB methods obtain PA-MPJPE of 56--107 mm, and lifting methods obtain 94--255 mm.
MotionAGFormer reaches 255 mm and 228 mm from input views~2 and~4, which are ID
view~2 and OOD view~2. GVHMR obtains 66 mm from input view~3, which is OOD
view~1, with a generated caption describing
jogging in place.

\end{document}